\documentclass[sigconf]{acmart}
 \PassOptionsToPackage{table}{xcolor}

\AtBeginDocument{%
  \providecommand\BibTeX{{%
    \normalfont B\kern-0.5em{\scshape i\kern-0.25em b}\kern-0.8em\TeX}}}

\usepackage{balance}
\copyrightyear{2023}
\acmYear{2023}
\setcopyright{acmlicensed}\acmConference[CIKM '23]{Proceedings of the 32nd
ACM International Conference on Information and Knowledge
Management}{October 21--25, 2023}{Birmingham, United Kingdom}
\acmBooktitle{Proceedings of the 32nd ACM International Conference on
Information and Knowledge Management (CIKM '23), October 21--25, 2023,
Birmingham, United Kingdom}
\acmPrice{15.00}
\acmDOI{10.1145/3583780.3615220}
\acmISBN{979-8-4007-0124-5/23/10}

\usepackage{hyperref}
\usepackage{url}
\usepackage{multirow}
\usepackage{graphicx}
\usepackage{float}
\usepackage{subfigure}
\usepackage{enumitem}
\usepackage{booktabs}
\usepackage{pifont}
\usepackage{caption}
\usepackage{xcolor,colortbl}
\usepackage{color}
\usepackage[utf8]{inputenc}
\usepackage{tcolorbox}
\usepackage{soul}

\begin{document}

\title[Unlocking the Potential of User Feedback: Leveraging LLMs as User Simulators to Enhance Dialogue]{Unlocking the Potential of User Feedback: Leveraging Large Language Model as User Simulator to Enhance Dialogue System}

\author{Zhiyuan Hu}
\email{zhiyuan\_hu@u.nus.edu}
\affiliation{%
  \institution{National University of Singapore}
      \country{}
  \postcode{}
}

\author{Yue Feng}
\email{yue.feng.20@ucl.ac.uk}
\affiliation{%
  \institution{University College London}
    \country{}
  \postcode{}
}

\author{Anh Tuan Luu}
\email{anhtuan.luu@ntu.edu.sg}
\affiliation{%
  \institution{Nanyang Technology University}
    \country{}
  \postcode{}
}

\author{Bryan Hooi}
\email{bhooi@comp.nus.edu.sg}
\affiliation{%
  \institution{National University of Singapore}
    \country{}
  \postcode{}
}

\author{Aldo Lipani}
\email{aldo.lipani@ucl.ac.uk}
\affiliation{%
  \institution{University College London}
    \country{}
  \postcode{}
}

\renewcommand{\shortauthors}{Zhiyuan Hu, Yue Feng, Anh Tuan Luu, Bryan Hooi, \& Aldo Lipani}

\begin{abstract}
Dialogue systems and large language models (LLMs) have gained considerable attention.
However, the direct utilization of LLMs as task-oriented dialogue (TOD) models has been found to underperform compared to smaller task-specific models. Nonetheless, it is crucial to acknowledge the significant potential of LLMs and explore improved approaches for leveraging their impressive abilities.
Motivated by the goal of leveraging LLMs, we propose an alternative approach called User-Guided Response Optimization (UGRO) to combine it with a smaller TOD model. This approach uses LLM as an annotation-free user simulator to assess dialogue responses, combining them with smaller fine-tuned end-to-end TOD models. By utilizing the satisfaction feedback generated by LLMs, UGRO further optimizes the 
supervised fine-tuned TOD model.
Specifically, the TOD model takes the dialogue history as input and, with the assistance of the user simulator's feedback, generates high-satisfaction responses that meet the user's requirements. 
Through empirical experiments on two TOD benchmarks, we validate the effectiveness of our method. The results demonstrate that our approach outperforms previous state-of-the-art (SOTA) results. 
\end{abstract}




\newtcolorbox{conversationbox}{
    colback=white,
    colframe=black,
    sharp corners,
    boxrule=1pt,
    width=0.47\textwidth
}



\maketitle

\section{Introduction}

In recent years, there has been significant interest in dialogue systems and their potential to improve human-computer interaction. The development of LLMs has particularly contributed to the advancement of general dialogue systems. However, adapting LLMs for task-oriented dialogues (TODs) in specific domains, such as service reservation, remains challenging. LLMs struggle to handle the intricacies of TODs as they require domain knowledge, background information, and context to generate appropriate responses. 
Recent research \cite{hudevcek2023llms, li2023guiding} has aimed to tackle this challenge using few-shot or zero-shot learning. However, even when adapting Alpaca-LoRA-7B or ChatGPT, the BLEU score and Success metric for TOD tasks remain significantly low compared with fine-tuned end-to-end models. An alternative option for fine-tuning LLMs is resource-intensive. Moreover, whether fine-tuning will result in significant improvements is not always guaranteed \cite{andreas2021comparative} and can vary depending on several factors, including the similarity between pretraining tasks and fine-tuning tasks, the model size, and the duration of the fine-tuning process. \footnote{Code is available at: \url{https://github.com/zhiyuanhubj/UGRO-CIMK23}.} \footnote{The major work was conducted when first author was in University College London.}

\begin{figure}
    \centering
    \vspace{-\topsep}
    \includegraphics[width=0.40\textwidth]{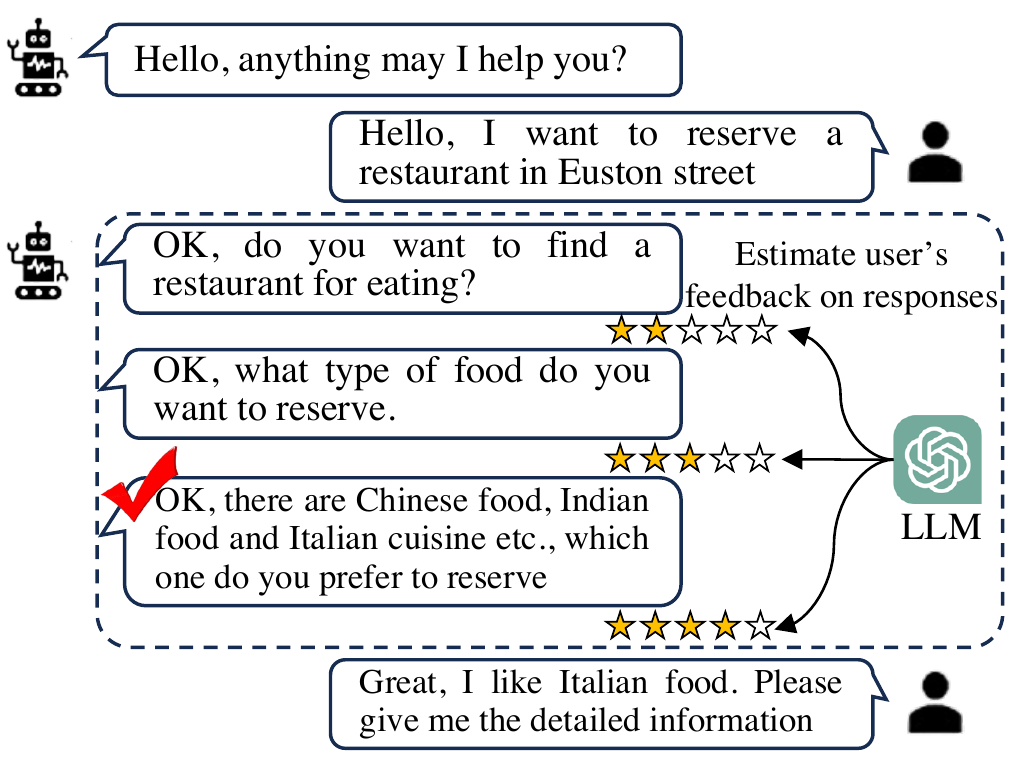}
    \vspace{-4mm}
    \caption{An example of using LLMs as a user simulator to predict user satisfaction scores in task-oriented dialogues.}
    \vspace{-6mm}
    \label{example_motivation}
\end{figure}

How should LLMs be integrated into TOD systems? Work in psychology has explored the dichotomy between two cognitive processes, System 1 (fast and intuitive) and System 2 (slow and deliberative), which guides or corrects System 1~\cite{kahneman2011thinking}, Our model design follows a similar separation of roles: we leverage the reasoning and knowledge of an LLM by using it as a source of \emph{feedback} to guide a task-specific and thus `intuitive' model.
Toward this goal, our approach is to utilize LLMs as a user simulator. 
User simulation has emerged as a line of work which involves mimicking user behaviour, including user query generation, system response satisfaction prediction, and user action prediction \cite{erbacher2022interactive, zerhoudi2022simiir, kim2022multi, sun2021simulating}. 
LLMs possess significant strengths, such as their remarkable understanding, extensive knowledge, and reasoning capabilities. Therefore, it is natural to utilize LLMs as user simulators to assess responses generated by TOD systems using smaller fine-tuning end-to-end models and provide feedback on user satisfaction to enhance the TOD system.

Figure~\ref{example_motivation} illustrates an example of a dialogue system response and a user simulator. In this scenario, the system responds to the user by considering the dialogue history. The user simulator, powered by an LLM, predicts the satisfaction scores based on the system response candidates and previous dialogue history. By leveraging these satisfaction feedback, we can select the most likely satisfying response for the user, thereby facilitating high-quality interactions. The key aspect enabling this is the ability of the LLM to assign a satisfaction score to each potential response candidate.

However, due to limitations in dataset annotation scale, modelling user behaviour, and the capabilities of current user simulator approaches, the performance of previous research on user simulators is not yet qualified for practical applications.
Consequently, we explore two questions: 1) whether current LLMs can serve as an effective user simulator and 2) how we can leverage the user simulator's feedback generated by LLMs to enhance the task-oriented dialogue system. To address these questions, 
we design suitable prompts and investigate the zero-shot and few-shot performance of LLM on user satisfaction prediction tasks. 
We propose our approach, User-Guided Response Optimization (UGRO), to leverage the user feedback to boost the TOD system.
Initially, we train the TOD model using a domain-supervised dataset.
Subsequently, the satisfaction score predicted by the LLM will be utilized as a reward to optimize the fully supervised trained TOD system. This will be accomplished by employing the Proximal Policy Optimization (PPO) algorithm, which allows us to capitalize on user feedback. In summary, our contributions can be categorized into three aspects:

\begin{itemize}[leftmargin=1em]
\item We investigate the performance of LLMs as a user simulator for providing feedback in zero-shot and few-shot settings.
\item We propose a new model called User-Guided Response Optimization (UGRO) to harness the potential of user feedback and enhance dialogue systems.
\item Extensive experiments validate the effectiveness of our model, including quantitative performance evaluation, human evaluation, and a case study.

\end{itemize}

\section{Related Work}

TOD systems facilitate tasks like hotel bookings or restaurant reservations. Some end-to-end models \cite{lee2021improving, he2022galaxy} generate responses using only the dialogue context, while policy optimization methods \cite{wang2020modelling, wang2020multi} use ground-truth dialogue states. 
\citet{lee2021improving, bang2023task} incorporate both text information and dialogue states to generate a dialogue response.
Additionally, reinforcement learning methods \cite{yu2022reinforced} have also shown improvements in TOD systems. \citet{hudevcek2023llms} show that specialized task-specific models still outperform general LLMs in TOD tasks. Additionally, the integration of LLMs into domain-specific dialogue systems has also been explored by \citet{snell2022context, li2023guiding}. 

Regarding user simulators, \citet{sun2021simulating} proposes the user satisfaction estimation task. It leverages human annotations regarding turn-level satisfaction to train an estimator. The estimator is then utilized to perform automatic evaluation by simulating users. \citet{deng2022user, kim2022multi} employ the multi-task framework to enhance the performance of their user simulator. \citet{feng2023schema} use schema-guided information, and \citet{ye2023modeling} incorporate satisfaction dynamics to enhance user satisfaction prediction. \cite{wei2023multi, ong2022discourse} explore 
Multi-Scale Receptive Field Graph model and  Variational-Autoencoder in unsupervised way to determine the emotion in a conversation.
 Additionally, \cite{li-etal-2022-controllable} propose a new dialogue simulation method based on LLM in-context learning to construct the dataset automatically. 


\section{Methodology}

\subsection{Overview}


We begin by assessing the performance of LLMs as user simulators for predicting response satisfaction. Subsequently, we fine-tune the TOD models through supervised training with response data and optimize it using the PPO algorithm and satisfaction rewards.

\vspace{-2mm}
\begin{figure*}[h]
\centering
    \vspace{-\topsep}
    \includegraphics[width=0.75\textwidth]{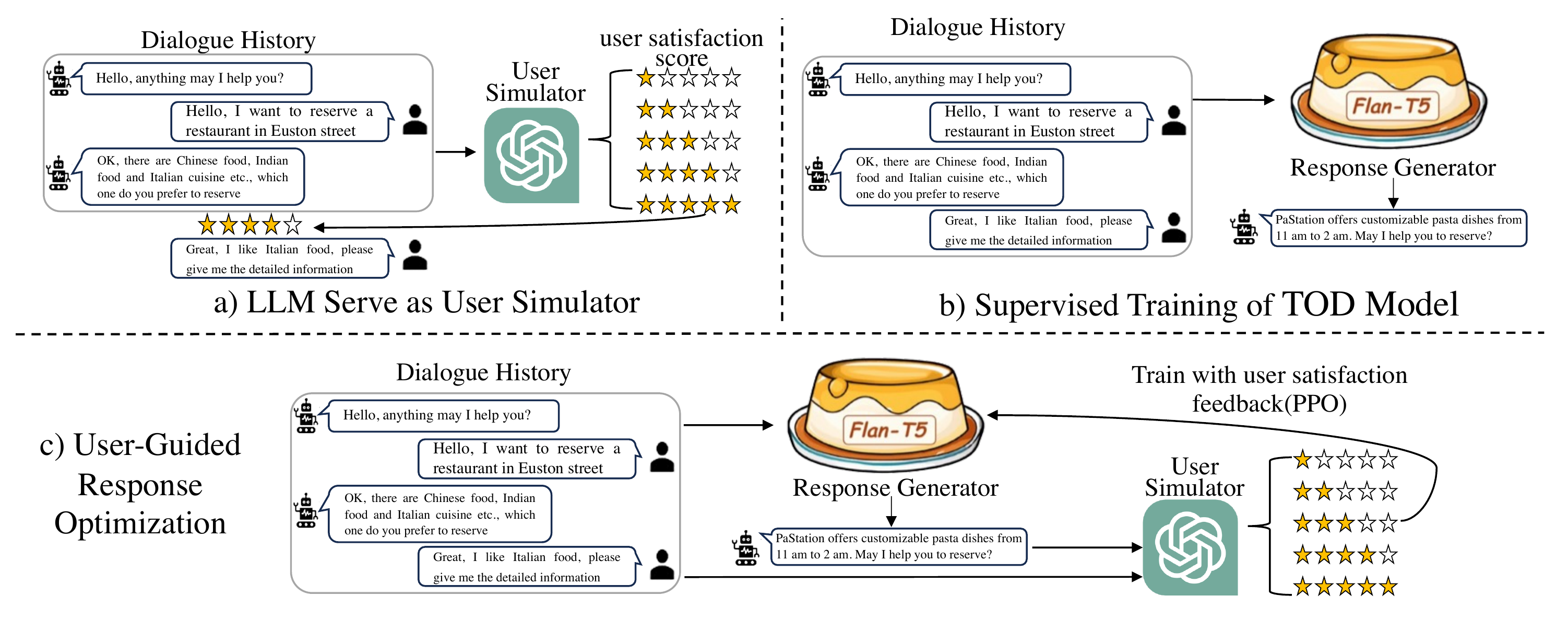}
    \vspace{-4mm}
    \caption{Model Architecture. TOD model undergoes supervised training initially and then is optimized based on user feedback.}
    \vspace{-4mm}
    \label{Model}
\end{figure*}

\subsection{LLM as User Simulator}


\textbf{Motivation}
\quad As mentioned earlier, the user simulation encompasses query generation, response satisfaction prediction, and action prediction. Satisfaction prediction stands out as the most direct feedback for assessing quality and enhancing the TOD model. 
Hence, in this work, our key hypothesis is that satisfaction prediction allows the capabilities of LLMs to be effectively \emph{transferred to a new domain} by prompting them in a suitable way to predict satisfaction scores in that domain.
LLMs' strong understanding ability and rich knowledge enable them to comprehend the semantic aspects of domain dialogue. Moreover, their reasoning capability allows them to provide a reasonable score with detailed explanations. 

As illustrated in Figure~\ref{Model}, when using LLMs as a user simulator, the dialogue history serves as the input, and the system response is considered as the last utterance. To provide the necessary context for LLM, the input prompt consists of the task description, criteria for satisfaction scores, data format introduction, 0-6 dialogue examples, the dialogue to be assessed, and instructions. Notably, we incorporate the idea of `Zero-shot Chain-of-Thought' \cite{kojima2022large} by including an `explanation reason' slot in the instruction. This allows the LLM to provide a reliable satisfaction score along with its reasoning. 
The complete prompt used in this work can be found in our GitHub repository.
The output of the user simulator consists of both the explanation reason and the satisfaction score. The score is extracted using regular expressions. 

Formally, the TOD dataset, denoted as $\mathcal{D}= \{\boldsymbol{x}, \boldsymbol{y}^*\}$, consists of dialogue history inputs $\boldsymbol{x}$ and corresponding ground truth response outputs $\boldsymbol{y}^*$. 
The TOD model's response is represented as $\boldsymbol{y}$, and the satisfaction score predicted by the LLM denoted as $\boldsymbol{S}$. 
\vspace{-0.1cm}
\begin{equation}
    \boldsymbol{S} = LLM(\boldsymbol{x}, \boldsymbol{y})
\end{equation}

\subsection{User-Guided Response Optimization}

\textbf{Using LLMs to enhance TOD models} \quad As discussed in our introduction, the limitations of directly utilizing LLMs in TOD systems necessitate a key design goal in our approach: employing LLMs as a user simulator to provide satisfaction feedback and leveraging this feedback to optimize the fine-tuning TOD model, so as to leverage the LLM's knowledge, understanding and reasoning capabilities.
This user feedback guides the TOD model toward producing responses that better satisfy the user.

In order to train a task-oriented dialogue (TOD) system, we begin by performing supervised training on the TOD model using the complete training dataset, maximizing the log-likelihood:
\vspace{-0.1cm}
\begin{equation}
   \mathcal{L}_{\mathrm{TOD}}=-\mathbb{E}_{\left(\boldsymbol{x}, \boldsymbol{y}^*\right) \sim \mathcal{D}} \log p_{\mathrm{TOD}}\left(\boldsymbol{y}^* \mid \boldsymbol{x}\right) 
\end{equation}

Our objective is to enhance the quality of TOD system-generated responses and improve user satisfaction. Thus, the optimization goal during this phase is to achieve higher satisfaction scores. Mathematically, we aim to maximize the following objective:
\vspace{-0.1cm}
\begin{equation}
    \mathbb{E}_{\boldsymbol{x} \sim \mathcal{D}, \boldsymbol{y} \sim p_{\mathrm{TOD}}}[\mathbb{R}(\boldsymbol{x},\boldsymbol{y})]
\end{equation}

where $\mathbb{R}(\boldsymbol{x}, \boldsymbol{y}) = \boldsymbol{S} = LLM(\boldsymbol{x}, \boldsymbol{y}), \boldsymbol{y} \sim p_{\mathrm{TOD}}(\cdot \mid \boldsymbol{x})$

The aforementioned optimization process can be complex and unstable for the TOD model. Therefore, we approach the TOD optimization as a reinforcement learning problem and employ the proximal policy optimization (PPO) algorithm \cite{schulman2017proximal}. We initialize the policy network $\pi_0$ using the TOD model, denoted as $p_{\mathrm{TOD}}$.
\vspace{-0.1cm}
\begin{equation}
\mathbb{E}_\pi[r]=\mathbb{E}_{\boldsymbol{x} \sim \mathcal{D}, \boldsymbol{y} \sim \pi(\cdot \mid \boldsymbol{x})}[r(\boldsymbol{x}, \boldsymbol{y})]    
\end{equation}

To incorporate penalty rewards, we also utilize the KL-divergence and set the hyperparameter $\beta$ for training. Consequently, the final reward is calculated as follows:
\vspace{-0.1cm}
\begin{equation}
r(\boldsymbol{x}, \boldsymbol{y})=LLM(\boldsymbol{x}, \boldsymbol{y})-\beta \log \frac{\pi(\boldsymbol{y} \mid \boldsymbol{x})}{p_{\mathrm{TOD}}(\boldsymbol{y} \mid \boldsymbol{x})}    
\end{equation}

\noindent \textbf{Implementation Details}  \quad In this work, ChatGPT 
is used as a user simulator to assess the generated response and provide satisfaction scores. We utilize Flan-T5 (large version) \cite{chung2022scaling}  as the fine-tuning TOD model. This model has been trained extensively on instruction tasks and is well-suited for domain-supervised fine-tuning.

\section{Experiment Results and Analysis}

\subsection{Experimental Settings}

\textbf{Dataset} \quad In dialogue response generation, MultiWoZ 2.1 \cite{budzianowski2018multiwoz} and Schema Guided Dialogue (SGD) \cite{rastogi2020towards} are two typical TOD datasets, and they also have been annotated with satisfaction scores.

\noindent \textbf{Evaluation Metrics} \quad For evaluation, we use BLEU  \cite{papineni2002bleu} and ROUGE (F1 score of ROUGE-1, ROUGE-2, and ROUGE-L) \cite{lin2004rouge} metrics to assess the generation quality and semantic performance.

\noindent \textbf{Compared Prior Art} \quad We compare our model with prior strong baselines, including HDNO \cite{wang2020modelling}, MTTOD \cite{lee2021improving}, PPTOD \cite{su2021multi}, GALAXY \cite{he2022galaxy}, and TOATOD \cite{bang2023task}. \textbf{HDNO} models the hierarchical structure between dialogue policy and natural language generation using an option framework \cite{} (temporal abstraction for reinforcement learning). \textbf{MTTOD} combines pretrained language models with a multi-task learning framework for end-to-end TOD modelling by utilizing span prediction as an auxiliary task. \textbf{PPTOD} is a unified plug-and-play model for TOD. It employs a multi-task pre-training strategy to learn primary TOD task completion skills from diverse dialogue corpora. \textbf{GALAXY} is a generative pretrained model that improves dialogue generation by incorporating semi-supervised learning and explicit policy injection. \textbf{TOATOD} is an end-to-end TOD system with task-specific adapters, learning independently for tasks such as dialogue state tracking and response generation. 


\begin{table*}[ht] 
\renewcommand{\arraystretch}{0.75}
\vspace{-3mm}
\caption{The user satisfaction prediction typically involves considering the user's utterance alongside the system response. However, in our case, user satisfaction is assessed by ChatGPT solely based on the system response, following the approach in \cite{sun2021simulating}, which is the SOTA performance. We use their BERT method as the baseline, replicating their experimental setup.}
\vspace{-4mm}
\centering
\begin{tabular}{cl|cccc|cccc} 
\toprule
     \multirow{2}{*}{\bf Setting}  & \multirow{2}{*}{\bf Model} & \multicolumn{4}{c|}{\bf MultiWoZ 2.1} & \multicolumn{4}{c}{\bf SGD} \\
    
     & &Acc & P & R & F1 & Acc &P & R & F1 \\
    \midrule
    \multirow{2}*{Supervised Training}     &HiGRU+ATTN  &79.7 &24.2  &24.0   &24.0 &70.9  &25.8  &26.2  &26.1  \\ 
    &BERT	&\bf{85.4} &\bf{31.6}  &\bf{26.9}  &\bf{27.1} &\bf{77.8} &\bf{28.9}  &\bf{26.9}  &\bf{27.0}  \\ 
    
    \midrule
    \multirow{2}*{Zero shot}  &BERT &14.8  &13.5  &20.3  &5.8  & 11.3 & 18.1 & 22.2 & 6.1\\
    &ChatGPT &61.1  &21.6  &19.1  &19.1   & 51.6 & 22.6 & \bf{26.6} & 19.4 \\
    \midrule
    
    \multirow{1}*{Few shot} 
    &ChatGPT  &\bf{78.1}	 & \bf{22.4} &\bf{20.8}   &\bf{21.1} &\bf{72.6}  &\bf{26.6}  & 23.1  & \bf{23.6}  \\ 
\bottomrule
\end{tabular}

\vspace{-2mm}
\label{user simulator}
\end{table*}

\begin{table*}[ht] 
\renewcommand{\arraystretch}{0.75}
\vspace{-1mm}
\caption{Performance of system response generation. The BLEU performances of the MultiWOZ 2.1 dataset are sourced from published papers, and ROUGE scores are calculated based on the generated responses uploaded to the leaderboard or provided by the authors.
In the case of the SGD dataset performance, * denotes results implemented using official source codes.}
\vspace{-4mm}
\centering
\begin{tabular}{l|ccccc|ccccc}
\toprule
    \multirow{2}{*}{\bf Model} & \multicolumn{5}{c|}{\bf MultiWoZ 2.1} & \multicolumn{5}{c}{\bf SGD} \\
    
    & BLEU & ROUGE & ROUGE-1 &ROUGE-2 &ROUGE-L & BLEU & ROUGE & ROUGE-1 &ROUGE-2 &ROUGE-L  \\
\midrule
    HDNO\cite{wang2020modelling}   &18.9 &28.2 &34.6   &17.0  &32.9  &- &-   &-  &-  &- \\
    MTTOD\cite{lee2021improving}  &21.0    &30.2   &36.9   &18.6  &35.0  &5.7$^*$ &18.6$^*$   &23.8$^*$  &9.2$^*$  &22.7$^*$ \\ 
    PPTOD\cite{su2021multi}    &19.2 &31.2 &37.8 &19.6 &36.3 &9.7$^*$ &22.0$^*$ &27.3$^*$ &12.6$^*$ &26.1$^*$ \\
    GALAXY\cite{he2022galaxy}   & 20.0 &30.2   &36.7 &18.7 &35.1  &-  &-  &- &- &-   \\ 
    TOATOD\cite{bang2023task}  &17.1 &29.1 &35.3   &18.0  &34.0  &- &- &-   &-  &-  \\
    UGRO\textbf{-}		 & 19.2 &24.5 &30.0 &14.3 &28.9    &19.8  &35.8  &42.2 &24.7 &40.4 \\ 
    UGRO  & \bf{34.3} &\bf{31.7}   &\bf{38.5} &\bf{20.1} &\bf{36.5}  & \bf{24.6}  &\bf{37.4}  &\bf{44.0} &\bf{26.4} &\bf{41.8}  \\ 
\bottomrule

\end{tabular}

\vspace{-3mm}
\label{UGRO results}
\end{table*}

\subsection{User Simulator Performance and Analysis}

Table~\ref{user simulator} demonstrates that ChatGPT performs well even in a zero-shot setting. Increasing the number of examples to 6 shows a significant improvement in accuracy. The F1 score also exhibits enhancement. Furthermore, ChatGPT's ability to predict satisfaction scores is comparable to that of fine-tuned BERT models and significantly superior to the zero-shot BERT model. We also noticed that the dataset imbalance affects satisfaction prediction in ChatGPT. Rating 3 dominates 90\% of samples, while ratings 1 and 5 have less than 1\%. This poses challenges for accurate satisfaction classification. 

Although ChatGPT's performance may only be comparable, its advantage lies in its practicality as a user simulator. This annotation-free approach addresses the issues of high annotation cost and human bias encountered in previous satisfaction score labelling. Furthermore, this method can be easily extended to new domains or tasks, even with minimal or zero examples, while still maintaining comparable and practical performance.



\subsection{Advancing TOD through User Feedback}


\vspace{-1mm}
Table~\ref{UGRO results} shows UGRO's superior performance compared to the previous SOTA model on the MultiWoZ 2.1 dataset. UGRO achieves a 13\% BLEU improvement over the MTTOD model and sets a new SOTA in ROUGE, surpassing the PPTOD model. To validate the effectiveness of user simulator feedback generated by LLMs, we also conduct an ablation study by performing UGRO\textbf{-} (Non-feedback UGRO) experiments. The results show a roughly 15\% increase in BLEU and an average 7\% improvement in ROUGE, highlighting the benefits derived from user feedback.

Furthermore, we implemented three baselines (MTTOD with the highest BLEU score, PPTOD with the highest ROUGE results and UGRO\textbf{-}) from the MultiWoZ 2.1 dataset into the SGD dataset. By comparing the performance of these baselines with the UGRO model, we can assess the generalization and stability of our method.

\vspace{-4mm}
\subsection{Case Study}

\begin{figure}
    \centering
    \vspace{-\topsep}
    \includegraphics[width=0.35\textwidth]{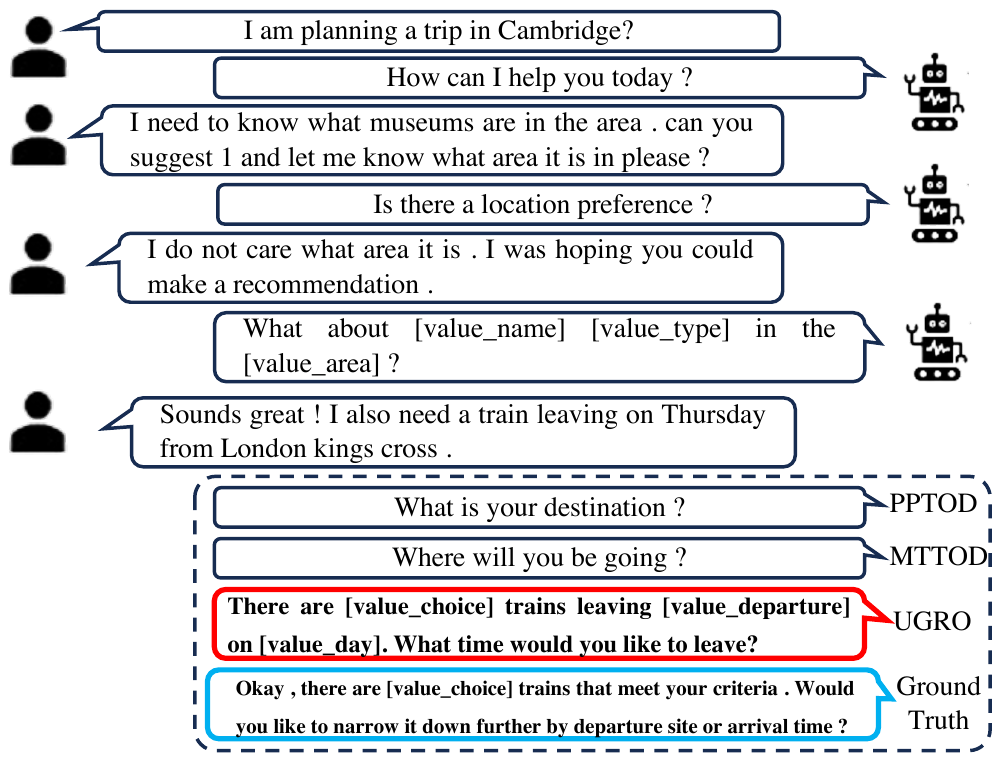}
    \vspace{-5mm}
    \caption{Dialogue responses generated by different models.}
    \vspace{-6mm}
    
    \label{example}
\end{figure}

\vspace{-1mm}
In Figure~\ref{example}, we compare the responses generated by our model UGRO, PPTOD and MTTOD to better illustrate our motivation. UGRO demonstrates a better understanding of dialogue history, providing clear and relevant responses while also requesting the necessary information. 
Furthermore, unlike the ground truth response that simply asks for the `departure' place, which has already been mentioned in the previous dialogue, the UGRO model provides both the departure site and the day for the train, resulting in a more satisfactory and higher-quality answer.


\vspace{-4mm}
\subsection{Human Evaluation}
\vspace{-1mm}
To ensure a more comprehensive evaluation, we incorporate human assessment to measure the satisfaction level and semantic quality of the generated responses. For satisfaction assessment, we employ a 5-level scoring setting with defined criteria in \cite{sun2021simulating}. Semantic quality is evaluated on a scale of 1 to 10, focusing on fluency and coherence. We randomly select 50 responses from each MTTOD, PPTOD, and UGRO generation results in the MultiWoZ 2.1 dataset and obtained assessments from 5 annotators.
Among the evaluated models, UGRO demonstrates superior performance, with a satisfaction score of 4.25, surpassing MTTOD (4.07) and PPTOD (4.09). Additionally, UGRO achieves the highest semantic quality score of 7.99, compared to MTTOD (7.89) and PPTOD (7.87), further solidifying its position as the top-performing model.



\section{Conclusion}

In this study, we embark on an exploration of leveraging LLMs as user simulators. We propose UGRO, a novel approach that optimizes the fine-tuned TOD model by incorporating satisfaction feedback from the LLM-powered user simulator. Our objective is to leverage the user simulator's potential and seamlessly integrate it into future TOD models. By using LLMs as user simulators with appropriate prompts, we achieve comparable performance to the previous SOTA model in predicting satisfaction scores. This annotation-free method can be extended to different domains. We validate UGRO by evaluating improvements in generated responses on the MultiWoZ 2.1 and SGD datasets, along with a case study and human evaluation that demonstrate enhancements in user satisfaction and semantic quality.
Looking ahead, we aim to further explore different forms of interaction between the LLM and TOD systems, such as explanatory reasons for satisfaction scores provided by LLM.



\bibliographystyle{ACM-Reference-Format}
\balance
\bibliography{sample-base}


\end{document}